\documentclass[runningheads]{llncs}

\usepackage{eccv}

\usepackage{eccvabbrv}

\usepackage{graphicx}
\usepackage{booktabs}

\usepackage[accsupp]{axessibility}  

\usepackage{multirow}
\usepackage{makecell} 

\usepackage{hyperref}

\usepackage{orcidlink}

\begin{document}
\title{Detect Fake with Fake: Leveraging Synthetic Data-driven Representation for Synthetic Image Detection} 

\titlerunning{Detect Fake with Fake}

\author{Hina Otake\thanks{First two authors contributed equally.}\inst{1,2}\orcidlink{0009-0000-3382-3079} \and
Yoshihiro Fukuhara$^{\star}$\inst{1,2}\orcidlink{0000-0001-8892-5339} \and
Yoshiki Kubotani\inst{2}\orcidlink{0000-0001-5448-7828} \and
Shigeo Morishima\inst{3}\orcidlink{0000-0001-8859-6539}}

\authorrunning{H.~Otake et al.}

\institute{Waseda University, Japan\\
\email{\{fumiwar88@akane,f\_yoshi@ruri\}.waseda.jp}
\and
cvpaper.challenge \\
\email{cvpaper.challenge@gmail.com}\\
\and
Waseda Research Institute for Science and Engineering, Japan \\
\email{shigeo@waseda.jp}}

\maketitle

\begin{abstract}
    Are general-purpose visual representations acquired solely from synthetic data useful for detecting fake images? In this work, we show the effectiveness of synthetic data-driven representations for synthetic image detection. Upon analysis, we find that vision transformers trained by the latest visual representation learners with synthetic data can effectively distinguish fake from real images without seeing any real images during pre-training. Notably, using SynCLR as the backbone in a state-of-the-art detection method demonstrates a performance improvement of $\boldsymbol{+10.32}$ mAP and $\boldsymbol{+4.73}\%$ accuracy over the widely used CLIP, when tested on previously unseen GAN models. Code is available at \url{https://github.com/cvpaperchallenge/detect-fake-with-fake}.
\end{abstract}

\section{Introduction}
\label{sec:intro}
With the societal widespread of generative models such as generative adversarial network (GAN) and diffusion model (DM), synthetic images have become easily accessible. In recent years, advancements in image generation technology have significantly reduced artifacts in synthetic images, making it more difficult to distinguish them from real images. Consequently, the misuse of synthetic images has led to serious social issues such as fake news, political and economic disruption, and identity fraud~\cite{cardenuto2023age, lin2024detecting}. As the quality of synthetic images improves, the impact of their misuse can no longer be ignored. Therefore, the development of methods for accurately detecting a wide variety of synthetic images has become a socially important mission to ensure the reliability of information.

In response to such societal demands, the scientific community has recently focused on developing methods for synthetic image detection (SID). Specifically, methods that use neural networks to learn the differences between real and fake images have been proposed~\cite{frank2020leveraging, marra2018detection, rossler2019faceforensics++, shiohara2022detecting}. However, approaches that explicitly train feature extractors for SID confront the problem of overfitting the types of fakes used during training, making it difficult to achieve strong generalization across different types of generative models~\cite{cozzolino2018forensictransfer, zhang2019detecting}. As a countermeasure, features extracted by foundation models like CLIP~\cite{radford2021learning} have been utilized~\cite{ojha2023towards, koutlis2024leveraging, liu2024forgery,liu2024mixture, cozzolino2024raising}. These methods leverage the general-purpose feature representations acquired by foundation models to achieve high generalization performance across various generative models.

Typically, these foundation models are pre-trained using large-scale datasets composed of real data. However, recent studies have proposed methods for training foundation models exclusively on synthetic data~\cite{takashima2023visual, NEURIPS2023_971f1e59, Tian_2024_CVPR}. These synthetic data-driven general-purpose representation learners achieve performance equal to or surpassing existing foundation models like CLIP and DINOv2~\cite{oquab2023dinov2} in tasks such as classification and segmentation~\cite{Singh_2024_CVPR}. This raises a fundamental question: "Are general-purpose feature representations learned solely from synthetic data, namely fake data, effective for SID?"

In this study, we evaluate the effectiveness of the synthetic data-driven general-purpose representations for SID using state-of-the-art methods such as StableRep~\cite{NEURIPS2023_971f1e59} and SynCLR~\cite{Tian_2024_CVPR}. Remarkably, we find that vision transformer (ViT)~\cite{dosovitskiy2020image} trained with StableRep and SynCLR acquired feature representations effective for distinguishing between fake and real images, despite never having seen real images during training. Moreover, SynCLR demonstrates superior performance to the widely used CLIP in detecting fakes generated by GANs and other generative models not used during pre-training.

Additionally, qualitative analysis suggests that universal representations derived from synthetic data capture different features than those learned from real images. Based on this analysis, we employ a simple ensemble learning approach and confirm that combining foundation models trained solely on synthetic data with ones on real data improves generalization performance for SID.


The contributions of this paper are threefold: (1) To the best of our knowledge, we are the first to analyze the effectiveness of using general-purpose feature representations trained exclusively on synthetic images as a backbone for SID. Our numerical evaluations across various datasets confirm that models with synthetic data-driven general-purpose representations outperform widely used baselines in detecting generative models not used during the backbone's pre-training phase. (2) We visualize the properties of the synthetic data-driven general-purpose representations. (3) We confirm that ensembling foundation models trained on real images with those trained on synthetic images effectively construct detectors with high generalization performance.

\section{Related Work}
\subsection{Synthetic Image}
Synthetic images come in various types, with deepfake being a prominent example. Deepfake technology utilizes deep learning to manipulate existing videos and audio, creating fictitious moving images that do not exist in reality. This technique primarily targets generating facial images of individuals. Deepfake generation methods are diverse, with face swapping being a representative technique. FaceShifter~\cite{li2019faceshifter} and SimSwap~\cite{chen2020simswap} create deepfakes by swapping the decoder of the trained GAN between the source and target images. Additionally, StyleSwap~\cite{xu2022styleswap} is a robust, high-quality face-swapping method that maps identity information into the latent space.

Other methods for creating deepfakes include expression swapping~\cite{thies2016face2face, Pumarola_2018_ECCV, nirkin2019fsgan} and attribute manipulation, which alter visual features like age, gender, and hair without changing an individual's unique identity~\cite{Choi_2018_CVPR, choi2020stargan}.
While these methods generate deepfakes based on real data, StyleGAN~\cite{karras2019style} and StyleGAN2~\cite{karras2020analyzing} utilize generative models to create entirely fictitious facial images. Those methods are frequently employed for entertainment but are often used for malicious intent.

Deepfakes primarily target facial images; however, recent advancements in GANs and DMs have facilitated the extensive replication of features and patterns present in natural images. Consequently, generating diverse and realistic images beyond the facial domain has become significantly more feasible. Notably, latent diffusion model (LDM)~\cite{Rombach_2022_CVPR} applies the diffusion process to the latent space rather than the pixel space, simultaneously improving the quality of synthetic images and reducing computational costs. By utilizing the powerful encoder of ViT trained with CLIP~\cite{radford2021learning} and the large-scale dataset LAION-5B~\cite{NEURIPS2022_a1859deb}, it has become possible to generate diverse, high-quality, and high-resolution images from text prompts. Additionally, DALL-E~\cite{ramesh2021zero, ramesh2022hierarchical, openai2023dalle3} uses a transformer as the encoder for VQ-VAE~\cite{oord2018neural, razavi2019generating} to create high-quality images from text. GigaGAN~\cite{kang2023gigagan}, with one billion parameters and cross-attention, generates images comparable to DMs and self-regressive models. Furthermore, methods such as Imagen~\cite{saharia2022photorealistic} and Midjourney~\cite{midjourney} specialize in generating high-resolution and photorealistic images, particularly excelling in the generation of complex scenes and diverse styles.

\subsection{Synthetic Image Detection}
With the advancement of powerful image generation and editing technologies, the need for techniques to detect such fake images has increased. Before the rapid development of generative models, methods were proposed to detect image manipulations by identifying anomalies such as abnormal reflections~\cite{10.1145/2077341.2077345}, resampling artifacts~\cite{popescu2005exposing}, and compression traces~\cite{agarwal2017photo}. Subsequently, with the development of deep learning and generative models such as GANs, the mainstream approach became training detectors that learn the artifacts~\cite{frank2020leveraging, zhang2019detecting} and inherent fingerprints~\cite{rossler2019faceforensics++,yu2019attributing,8695364} produced by generative models.

However, detectors that directly learn the features of synthetic images have been found to frequently overfit and fail to generalize across different types of generative models~\cite{cozzolino2018forensictransfer, zhang2019detecting}. To address this overfitting issue, various attempts have been made to improve generalization performance, including the use of carefully designed data augmentation~\cite{wang2020cnn, yan2024transcending}, metric learning~\cite{luo2023prior}, adversarial training in latent space~\cite{chen2022self}, detection of artifacts during upsampling~\cite{tan2024rethinking}, and formulation as a multi-class classification problem~\cite{shahid2024generalized}.

In these efforts to improve generalization performance, a method has been proposed to use the general-purpose feature representations acquired by CLIP~\cite{radford2021learning} directly for SID~\cite{ojha2023towards}, achieving significant performance improvements over previous baselines. Subsequent lines of work include methods such as using only the shallow layer features of CLIP~\cite{koutlis2024leveraging}, incorporating multiple LoRA modules into the CLIP encoder~\cite{liu2024mixture}, aligning CLIP's feature representations with text prompts~\cite{liu2024forgery}, and using backbone that combine multiple foundation models through ensemble learning~\cite{azizpour2024e3, nguyen2024exploring} or MLP-Mixer~\cite{ESSA2024128128}. Similar to these works, this study also employs the general-purpose representations acquired by foundation models for SID. However, while previous studies have been limited to analyzing foundation models trained on real data, such as CLIP, we aim to evaluate the effectiveness of feature representations from foundation models trained exclusively on synthetic data.

\subsection{Foundation Models Trained by Synthetic Data}
Foundation models are designed to acquire general-purpose representations effective for various downstream tasks. 
Examples include CLIP~\cite{radford2021learning}, which is trained on text-image pairs, DINO~\cite{caron2021emerging, oquab2023dinov2}, which uses self-distillation without labels, and 4M~\cite{4m, 4m21}, which is based on multimodal training. These foundation models are typically pre-trained using large-scale datasets on the scale of millions or billions of real data. However, the creation and cleansing of such datasets incur significant costs.

In response to the challenges of constructing such large-scale datasets, methods for acquiring general-purpose feature representations using synthetic data have been proposed. Pioneering research includes methods that generate training data based on mathematical rules~\cite{KataokaACCV2020, 9880154, Nakashima_Kataoka_Matsumoto_Iwata_Inoue_Satoh_2022,  takashima2023visual} such as fractals or circular harmonics, use random tiling images of various shapes for training~\cite{baradad2021learning}, or employ geometrical images generated from programming code~\cite{baradad2022procedural}. However, the performance of these models has not reached the level of powerful foundation models like CLIP.

More recently, methods for training foundation models using data synthesized by generative models have been proposed. StableRep~\cite{NEURIPS2023_971f1e59} uses images generated by DMs from captions of real image datasets for training. Subsequently, SynCLR~\cite{Tian_2024_CVPR} takes this further by using text generated by language models as input to DMs. These methods have achieved performance comparable to or surpassing those trained on real data, such as CLIP. It has been reported that the synthetic data-driven general-purpose representations acquired by these foundation models possess different properties from those learned from real data~\cite{Singh_2024_CVPR, Fan_2024_CVPR}, though many aspects remain unclear. In this study, we evaluate the effectiveness of synthetic data-driven representations in SID.
\section{Preliminaries}

\subsection{Problem Setup}

Let $\mathcal{X}\subset\mathbb{R}^{d}$ denotes the input space, where $d$ is the data dimension. SID is a task that classifies whether a given image $\boldsymbol{x}\in\mathcal{X}$ was naturally captured using a camera (real) or is a synthetic image (fake). In this study, we define synthetic images as those artificially generated or edited using generative models. The current major paradigm for this task involves training a neural network as binary classifier $f: \mathcal{X}\to\mathbb{R}$, which outputs a label indicating whether the input image is real (0) or fake (1).


\subsection{UnivFD}

Features extracted by pre-trained foundation models have been shown to be remarkably effective for SID~\cite{ojha2023towards, koutlis2024leveraging, liu2024mixture, liu2024forgery, nguyen2024exploring, ESSA2024128128}. The use of powerful feature representations acquired by foundation models mitigates the issue of overfitting to the generative models used during training. This results in high generalization performance across diverse generative models.

UnivFD~\cite{ojha2023towards} is the first method to employ this approach. In UnivFD, the parameters of the feature extractor $\phi:\mathbb{R}^{d}\to\mathbb{R}^{n}$ are frozen, where $n$ is the embedding space dimension.
The parameters of the detector, denoted as $\boldsymbol{\theta}$, are trained using binary cross-entropy (BCE) loss, as shown in Equation (\ref{loss}):

\begin{equation}
\label{loss}
\mathcal{L} = -\sum_{\boldsymbol{x}\in\mathcal{F}} \mathrm{log}\Big[\psi_{\boldsymbol{\theta}}(\phi(\boldsymbol{x}))\Big] - \sum_{\boldsymbol{x}\in\mathcal{R}} \mathrm{log}\Big[1 - \psi_{\boldsymbol{\theta}}(\phi(\boldsymbol{x}))\Big]
\end{equation}
Here, $\psi_{\boldsymbol{\theta}}:\mathbb{R}^{n}\to\mathbb{R}$ is a single fully connected layer with a sigmoid activation function, and $\mathcal{R}$ and $\mathcal{F}$ are the sets of real images and fake images in the training data, respectively. Additionally, a ViT~\cite{dosovitskiy2020image} pre-trained with CLIP~\cite{radford2021learning} is employed as the foundation model for $\phi$. In our experiments, we also adopt UnivFD as the synthetic image detector, but for $\phi$, we use ViTs trained by various methods including CLIP.

\section{Experiments}
\begin{table}[tb]
    \caption{Comparison of pre-training conditions and backbones of foundation models.}
    \label{table:condition}
    \centering
    \begin{tabular}{lcccc}
        \toprule
        & text & image & $\#$ images & backbone \\
        \midrule
        CLIP~\cite{radford2021learning}& real & real & 400M & ViT-B/16 \\
        DINOv2~\cite{oquab2023dinov2}& - & real & 142M & ViT-B/14 \\
        \midrule
        StableRep~\cite{NEURIPS2023_971f1e59} & real & syn & 100M & ViT-B/16 \\
        SynCLR~\cite{Tian_2024_CVPR}& syn & syn & 600M & ViT-B/16 \\
        \bottomrule
    \end{tabular}
\end{table}
In all experiments, we use UnivFD as the framework for SID. Different pre-trained foundation models are adopted as the backbone of UnivFD, and we analyze their impact. We use ViT-B a variant of ViT, as the architecture for the backbone. For pre-training the backbone, we employ CLIP and DINOv2 with real images, and StableRep and SynCLR with synthetic data. A comparison of the pre-training conditions is shown in Table~\ref{table:condition}. 

To use publicly available weights, we adopt CLIP trained on LAION-400M~\cite{schuhmann2021laion} as published in OpenCLIP~\cite{ilharco_gabriel_2021_5143773}.
Similar to the original UnivFD paper, we use only ProGAN’s~\cite{karras2017progressive} training data to train the fully connected layer.
The optimization methods and hyperparameters for training are also set in the same way as in the original paper.

\subsection{How Useful are General-purpose Synthetic Data-driven Representations for SID?}
\label{subsec:how_useful}

To analyze the impact of the feature representations learned by the backbone on detection performance, we follow previous work~\cite{ojha2023towards, wang2020cnn} and evaluate performance against generative models. These include GAN-based methods such as ProGAN~\cite{karras2017progressive}, CycleGAN~\cite{zhu2017unpaired}, Big-GAN~\cite{brock2018large},  StyleGAN2~\cite{karras2020analyzing}, StarGAN~\cite{Choi_2018_CVPR}, and GauGAN~\cite{park2019semantic}, GigaGAN~\cite{kang2023gigagan}, as well as DM-based methods including the Guided Diffusion Model~\cite{dhariwal2021diffusion}, LDM~\cite{rombach2022high}, and Glide~\cite{nichol2021glide}. For the LDM, we generated images using 200 steps of denoising, with and without classifier-free guidance (CFG). The pre-trained Glide model used 100 steps to initially upsample an image to 64$\times$64, then employed an additional 27 or 10 steps to achieve a final resolution of 256$\times$256. Additionally, we evaluate on other methods such as DeepFakes~\cite{rossler2019faceforensics++}, SITD~\cite{chen2018learning}, SAN~\cite{dai2019second}, CRN~\cite{chen2017photographic}, IMLE~\cite{li2019diverse}, and DALL-E~\cite{ramesh2021zero}. Each generative model has a collection of real and fake images. As evaluation metrics, We follow existing work and report both average precision (AP) and classification accuracy.

While the primary purpose of this experiment is to compare the impact of feature representations learned by different backbones on detection performance using the UnivFD framework, we also include the results of two state-of-the-art SID methods as a performance reference:

\begin{enumerate}
  \item Wang~\cite{wang2020cnn}: A standard ResNet-50~\cite{he2016deep} architecture, pre-trained on ImageNet, is fine-tuned on SID with carefully chosen pre- and post-processing techniques, as well as data augmentations.
  \item LGrad~\cite{tan2023learning}: Image gradients, derived from a pre-trained deep neural network, are input into a standard ResNet-50 pre-trained on ImageNet which is fine-tuned for SID.
\end{enumerate}

\begin{table}[tb]
    \caption{Average precision (AP) for all backbone pre-training methods (rows) in detecting fake images from different generative models (columns). We note that the variant within Ours which uses CLIP as the backbone becomes identical configuration to the original UnivFD.
    }
    \label{table:ap}
    \centering
    \resizebox{\textwidth}{!}{%
    \begin{tabular}{clp{3em}p{3em}p{3em}p{3.5em}p{3em}p{3em}p{3em}p{3em}p{3em}p{3em}p{3em}p{3em}p{3.5em}p{3em}p{3.5em}p{3.5em}p{3.5em}p{4em}p{3em}} 
    \toprule
        \multirow{2}{*}{Method} &
        \multirow{2}{*}{Variant} &
        \multicolumn{7}{c}{Generative Adversarial Networks} & 
        \multirow{2}{*}{\makecell[tl]{Deep\\fakes}}&
        \multicolumn{2}{c}{Low vision} &
        \multicolumn{2}{c}{Perceptual} &
        \multirow{2}{*}{Guided} &
        \multicolumn{2}{c}{LDM} &
        \multicolumn{2}{c}{Glide} &
        \multirow{2}{*}{\makecell[tl]{DALL-E}}  &
        \multicolumn{1}{c}{Total}
        \\  \cmidrule(r){3-9} \cmidrule(r){11-14} \cmidrule(r){16-19} \cmidrule{21-21}
        &&Pro- GAN & Cycle- GAN & Big- GAN & Style- GAN2 & Gau- GAN & Star- GAN &Giga- GAN& & SITD &  SAN & CRN & IMLE & & 200 steps & 200 w/CFG & 100 27 & 100 10 & & mAP
        \\
        \midrule
        \multirow{2}{*}{Wang~\cite{wang2020cnn}} &prob. 0.5&99.98& 94.78 & 85.07&83.53& 97.01&95.12&57.25&72.29 &92.10 &59.87&\textbf{98.97}&\textbf{99.56}& 70.15& 75.46&76.88& 73.35& 80.76 &81.37&82.97\\
        &prob. 0.1&\textbf{100.0}& 89.63 & 82.21&86.92& 87.16&98.00&61.78&\textbf{91.57}&92.91 &68.57 &97.62&98.19&77.75& 74.75& 74.75&85.25& 86.87& 82.23 &85.34\\
        \midrule
        \multirow{3}{*}{LGrad~\cite{tan2023learning}} & 1-class &99.88& 90.56 & 84.73& 66.72& 76.03&99.81&74.40&88.13&59.41 &54.55 &81.56&80.93&75.08& 95.22& 96.51&90.11& 92.18& 95.70 &83.42\\
        & 2-class &99.99& 92.80 & 90.28& 68.52& 76.36&\textbf{99.98}&76.40&75.31&65.93 &56.37 &59.33&80.44&80.24& 96.15& 97.20&\textbf{94.63}& \textbf{95.82}& 95.31 &83.39\\
        & 4-class &99.99& 91.72 & 85.97& 73.81& 71.62&99.95&\textbf{79.99}&76.47&55.98 &59.48 &60.49&66.82&83.94& \textbf{98.25}& \textbf{98.59}&93.06& 94.94& \textbf{95.81} &82.60\\
        \midrule
        \multirow{4}{*}{Ours} & CLIP(UnivFD~\cite{ojha2023towards})&99.91& 93.40 & 88.03&62.17& 96.90&93.60&62.01&80.55 &77.98 &65.55&75.66&97.91& \textbf{89.64}& 92.87&76.88& 86.26& 85.40 &89.94&84.15\\
        & DINOv2 &99.82 & 93.90& 94.93 &68.74 &98.76&94.08&74.21 &75.71& 91.80& 71.05&74.57&86.65 &82.08& 96.05 & 83.04 &90.82&89.25 & 89.31&86.38\\
        & StableRep & 99.93 & 90.56 & 85.00&83.64&98.24&85.85 &63.36&86.33&96.70 &70.44&91.59& 96.01 & 64.64 & 87.22& 66.91& 75.06 &74.90 & 70.80 &82.62\\
        & SynCLR &99.97 & \textbf{97.03} & \textbf{98.25}&\textbf{90.75}&\textbf{99.92}&96.75&75.34&80.19 &\textbf{99.84} &\textbf{79.34}&98.66&99.50& 71.65 & 92.01& 78.19& 85.64 & 85.02 & 87.97 &\textbf{89.78}\\
        \bottomrule
    \end{tabular}
    }
\end{table}
\begin{table}[tb]
    \caption{Classification accuracy for all backbone pre-training methods (rows) averaged over real and fake classes for each generative model (columns). We note that the variant within Ours which uses CLIP as the backbone becomes identical configuration to the original UnivFD.}
    \label{table:acc}
    \centering
    \resizebox{\textwidth}{!}{%
    \begin{tabular}{clp{3em}p{3em}p{3em}p{3.5em}p{3em}p{3em}p{3em}p{3em}p{3em}p{3em}p{3em}p{3em}p{3.5em}p{3em}p{3.5em}p{3.5em}p{3.5em}p{4em}p{3em}} 
    \toprule
        \multirow{2}{*}{Model} &
        \multirow{2}{*}{Variant} &
        \multicolumn{7}{c}{Generative Adversarial Networks} & 
        \multirow{2}{*}{\makecell[tl]{Deep\\fakes}}&
        \multicolumn{2}{c}{Low vision} &
        \multicolumn{2}{c}{Perceptual} &
        \multirow{2}{*}{Guided} &
        \multicolumn{2}{c}{LDM} &
        \multicolumn{2}{c}{Glide} &
        \multirow{2}{*}{\makecell[tl]{DALL-E}}  &
        \multicolumn{1}{c}{Total}
        \\ \cmidrule(r){3-9} \cmidrule(r){11-14} \cmidrule(r){16-19}  \cmidrule{21-21}
        &&Pro- GAN & Cycle- GAN & Big- GAN & Style- GAN2 & Gau- GAN & Star- GAN &Giga- GAN& & SITD &  SAN & CRN & IMLE & & 200 steps & 200 w/CFG & 100 27 & 100 10 & & Avg. acc
        \\
        \midrule
        \multirow{2}{*}{Wang~\cite{wang2020cnn}} &prob. 0.5&99.20& 75.30 & 56.25&60.05& 76.30&74.95&50.80&52.20&79.50&50.00 &85.10&92.85&52.75&50.20& 50.25& 51.40&51.90& 52.25& 64.51 \\
        &prob. 0.1&\textbf{99.90}& 83.05 & 69.00&\textbf{79.45}& 77.40&90.65&54.65&55.70&87.00&51.50 &\textbf{87.15}&87.20&62.70&52.05& 52.25& 58.40&59.70& 57.10& 70.27 \\
        \midrule
        \multirow{3}{*}{LGrad~\cite{tan2023learning}} & 1-class &98.50& 81.75 & 78.35& 63.45& 71.00&97.70&67.65&74.45&60.50 &51.00 &53.85&54.10&67.95& 85.85& 88.75&80.80& 83.10& \textbf{87.15} &74.77\\
        & 2-class &99.40& 84.80 & 80.60& 62.05& 71.45&\textbf{99.55}&71.05&66.85&58.00 &56.00 &52.35&53.15&71.00& 89.40& 90.90&\textbf{86.80}& 88.65& 86.60 &\textbf{76.03}\\
        & 4-class &99.65& 82.40 & 79.05& 62.25& 69.00&98.60&\textbf{73.15}&63.80&57.50 &\textbf{59.00}&50.80&50.80&75.60&\textbf{92.10}& \textbf{93.65}&86.35& \textbf{88.75}& 85.30 &75.99\\
        \midrule
        \multirow{4}{*}{Ours} &CLIP(UnivFD~\cite{ojha2023towards})&98.40& 84.45 & 80.10&58.05& 89.30&84.40& 56.35&73.00 &66.00 &57.50&63.45&\textbf{93.05}& \textbf{81.95}& 82.20&62.55& 71.95& 70.15 & 77.15 &75.00\\
        &DINOv2&98.20 & 85.40& 85.00 &57.90 &92.90&82.00&63.15&65.15& 75.50& 58.50&52.85&59.40 & 69.40&89.70 &69.95&79.65&78.75& 77.65&74.50\\
        &StableRep& 98.75 & 78.25 & 63.75&72.60&87.50&70.60 &53.20&\textbf{76.65}&77.50&54.00&55.45&60.15&53.20&69.20 &53.80&58.20& 58.40 & 56.50& 66.54 \\
        &SynCLR &99.55 & \textbf{90.30} & \textbf{91.70}&55.65&\textbf{98.35}&87.70&57.30&73.10 &\textbf{96.00}&57.00&69.50&82.30 & 53.45 & 68.95& 56.15& 62.00 & 61.45 & 65.30 &73.65\\
        \bottomrule
    \end{tabular}
    }
\end{table}

Table~\ref{table:ap} and Table~\ref{table:acc} show AP and classification accuracy, respectively, of all backbone pre-training methods (rows) in detecting fake images from different generative models (columns). For classification accuracy, the numbers shown are averaged over the real and fake classes for each generative model.

The numerical results indicate that ViTs trained with StableRep and SynCLR, which acquire synthetic data-driven representation, can distinguish between real and fake images, even for fakes unseen during the training of their detectors. Remarkably, despite these foundational models never being exposed to GAN-generated or real images during their pre-training, StableRep and SynCLR demonstrate high detection performance for images from the GAN family. Specifically, SynCLR improves by $+$10.32 mAP and $+$4.73\% accuracy on average compared to CLIP within the GAN family.

In contrast, the detection performance is generally low for images generated by the DM family, which were used during pre-training. 
The reason for this could be that during the pre-training of StableRep and SynCLR, all images contain artifacts originating from DMs. Consequently, the ability to capture these artifacts is of little use in solving the pre-training task, and it is likely that the models do not acquire representations that capture the characteristics of artifacts originating from DMs.

\begin{figure}[t]
    \centering
    \includegraphics[width=1.0\textwidth]{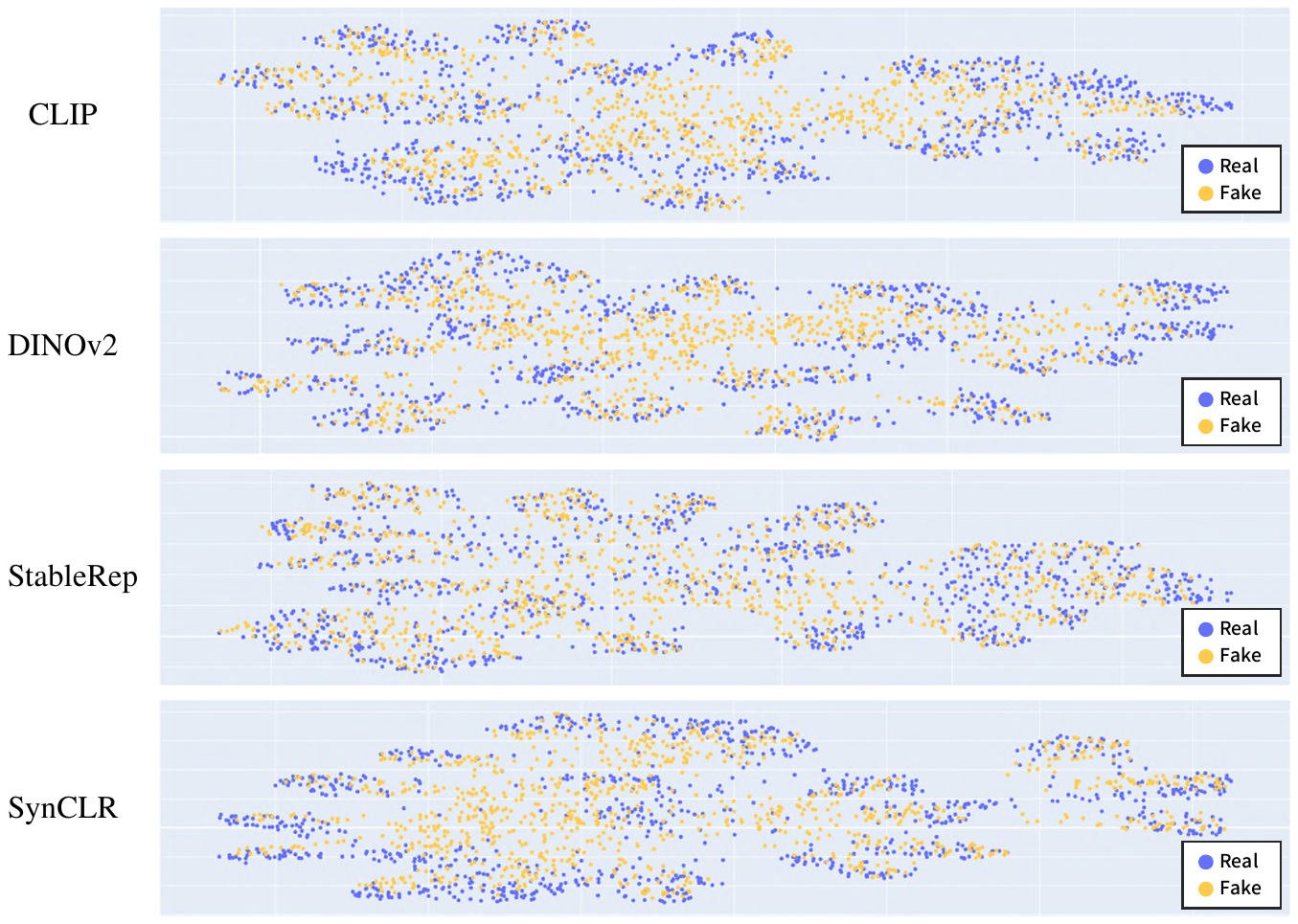}
    \caption{
        UMAP visualization of real images (blue) and fake images generated by ProGAN (yellow) in the backbone embedding space. SynCLR’s embedding space best separates the real features from fake.
    }
    \label{fig:feature}
\end{figure}
\subsection{Visual Analysis of Synthetic Data-driven Representations}
\begin{figure}[t]
    \centering
    \includegraphics[width=1.0\textwidth]{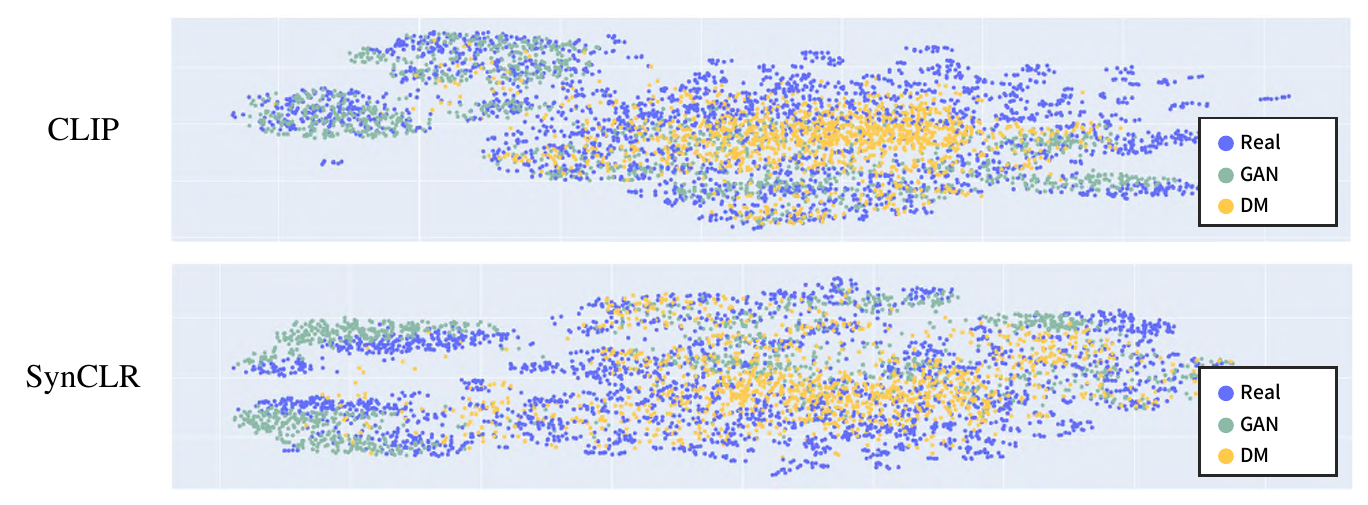}
    \caption{UMAP visualization of real images (blue), fake images generated by GANs (green), and fake images generated by DMs (yellow) using different backbone embedding spaces. The GAN data points include images generated by ProGAN, CycleGAN, BigGAN, StarGAN, and StyleGAN2. The DM data points include images generated by Guided, LDM, and Glide.}
    \label{fig:gan_dm_feature}
\end{figure}

So far, we have seen the surprisingly good performance of synthetic data-driven representations as a backbone for SID. In this section, we analyze the properties of synthetic data-driven representations using multiple visualization methods.

Fig.~\ref{fig:feature} shows a visual analysis of the embedding spaces of backbones pre-trained using different methods. Using the feature vectors from each model, we plotted four feature banks consisting of the same real and fake images obtained from ProGAN, and color-coded the resulting UMAP~\cite{mcinnes2018umap} plots with binary (real/fake) labels. All backbones exhibit a certain level of performance in separating real (blue) and fake (yellow) features, but the embedding space of SynCLR demonstrates the best separation performance.

We also provide visualizations to confirm the differences in detection performance across different types of generative models. Fig.~\ref{fig:gan_dm_feature}  shows a visualization of the embedding spaces using UMAP, similar to Fig.~\ref{fig:feature}, but the fake data includes images generated by various generative models. The GAN category includes images generated by ProGAN, CycleGAN, BigGAN, StarGAN, and StyleGAN2, while the DM category includes images generated by Guided, LDM, and Glide. The embedding space of SynCLR separates GAN and real images better compared to that of CLIP. On the other hand, the embedding space of SynCLR does not sufficiently separate DM and real images. This visualization result is consistent with the numerical evaluation results presented in the previous section.

\begin{figure}[p]
    \centering
    \includegraphics[width=0.9\textwidth]{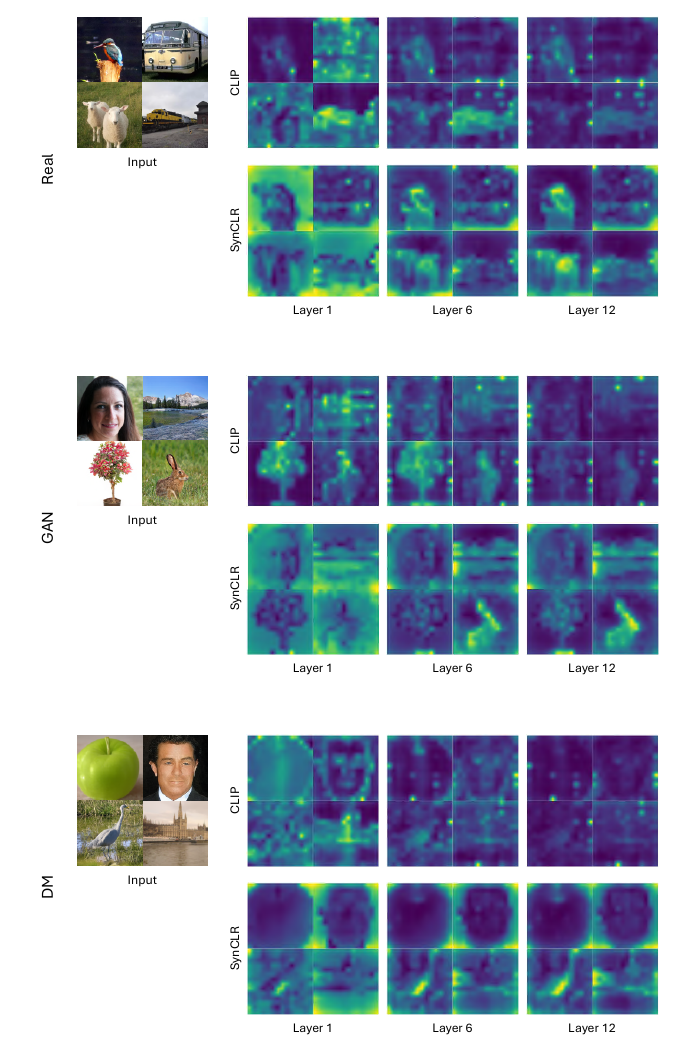}
    \caption{Attention maps visualizing the areas of focus for each model during SID. The maps show the first, intermediate, and last layers for real images and images generated by GANs (ProGAN, CycleGAN, BigGAN, StyleGAN2) and DMs (Guided, LDM, Glide), averaged across all heads.}
    \label{fig:attention_map}
\end{figure}
We use attention maps to visualize which parts of the images the backbones with synthetic data-driven representations are focusing on. Fig.~\ref{fig:attention_map} shows the results for CLIP and SynCLR for real and fake images. The attention maps are visualized for the initial layer, middle layer, and final layer, and the maps are averaged across all heads. The synthetic images used as sample inputs were generated by ProGAN, CycleGAN, BigGAN, and StyleGAN2 for GANs, and by Guided, LDM, and Glide for DMs. Compared to CLIP, SynCLR’s shallow layer maps show a broad attention spread across the entire image. As the layers deepen, there is a tendency for attention to focus more on the main elements. Additionally, in SynCLR’s maps, there are almost no artifacts caused by high-norm tokens~\cite{darcet2023vision} that are observed in CLIP’s maps, despite using the same ViT architecture. These observations qualitatively suggest that the synthetic data-driven representations acquired by SynCLR are highly different from those learned by CLIP.

\subsection{Evaluating the Effectiveness of Ensemble Learning with Synthetic Data-driven Representations}

In the previous section, qualitative analysis confirmed that synthetic data-driven representations capture different features compared to those learned using only real images. Based on this analysis, we conduct a simple ensemble learning experiment to verify whether combining backbones having synthetic data-driven representations with those having different representations can improve the performance of synthetic image detectors. For the ensemble method, we adopt feature fusion~\cite{hl2024multimodal,tu2022alzheimer,sun2005new,liu2001shape}, where the features are combined just before the fully connected layer, and the parameters of the fully connected layer are then trained using the combined features. Apart from adopting feature fusion, the training process is the same as described in Section~\ref{subsec:how_useful}.

\begin{table}[tb]
    \caption{Average precision (AP) for all combinations of backbone pre-training methods (rows) in detecting fake images from different generative models (columns). CLIP is the 32nd epoch of OpenCLIP, and $\text{CLIP}^{*}$ represents the weights of the 31st epoch. The combination of CLIP and $\text{CLIP}^{*}$ in the first row is the baseline.}
    \label{table:ap_ensemble}
    \centering
    \resizebox{\textwidth}{!}{%
    \begin{tabular}{ccp{3em}p{3em}p{3em}p{3.5em}p{3em}p{3em}p{3em}p{3em}p{3em}p{3em}p{3em}p{3em}p{3.5em}p{3em}p{3em}p{3.5em}p{3.5em}p{4em}p{3em}} 
    \toprule
        \multirow{2}{*}{Model} &
        \multicolumn{8}{c}{Generative Adversarial Networks} & 
        \multirow{2}{*}{\makecell[tl]{Deep\\fakes}} &
        \multicolumn{2}{c}{Low vision} &
        \multicolumn{2}{c}{Perceptual} &
        \multirow{2}{*}{Guided} &
        \multicolumn{2}{c}{LDM} &
        \multicolumn{2}{c}{Glide} &
        \multirow{2}{*}{\makecell[tl]{DALL-E}} &
        \multicolumn{1}{c}{Total}
        \\ \cmidrule(r){3-9} \cmidrule(r){11-14} \cmidrule{16-17} \cmidrule(r){18-19} \cmidrule{21-21} 
        & & Pro- GAN & Cycle- GAN & Big- GAN & Style- GAN2 & Gau- GAN & Star- GAN & Giga- GAN & & SITD &  SAN & CRN &  IMLE & & 200 steps & 200 w/CFG & 100 27 & 100 10 & & mAP
        \\
        \midrule
        CLIP & $\text{CLIP}^{*}$ &99.91& 93.22 & 88.31&62.77& 96.94&93.33&62.74&80.52 &79.53 &65.60 &75.20&97.87&89.99& 93.10& 77.69&87.15& 86.41& 90.35 &84.48\\
        \midrule
        CLIP & DINOv2  &\textbf{100.0} & 95.20& 98.07 &72.32 &99.68&97.00&79.14&84.20 &94.32& 68.63& 79.99&96.05&\textbf{92.18} & 96.42& 83.65 & 87.44 &86.12&90.90& 88.96 \\
        CLIP & StableRep & 99.99 & 92.68 & 91.16&79.10&99.12&92.24 &68.48&\textbf{88.51}&96.27&70.72 &89.96&98.01& 78.40 & 94.10 & 79.63& 84.45& 84.27 & 84.05 &87.29 \\
        CLIP & SynCLR &99.99 &\textbf{96.70}  & 98.60&90.44&\textbf{99.97}&\textbf{97.67}&77.44&84.76&\textbf{99.66}&80.44 &98.06&\textbf{99.81}&80.41& 95.51 & 84.36& 90.37 & 89.62 &92.30&  92.01\\
        DINOv2&StableRep & \textbf{100.0}& 94.32& 97.28&80.85&99.61 &93.86&79.05&86.24&95.49 &73.40 &90.73&95.73& 83.97& 96.71 & \textbf{96.84} & 89.93& 89.25 &87.82 & 90.62 \\
        DINOv2 &SynCLR &\textbf{100.0} & 96.44 &\textbf{99.30}&88.74&99.94 &96.38 &\textbf{79.76}&82.37 &97.16 &75.96&96.67&99.19& 82.78 & \textbf{97.21} & 87.27 & \textbf{93.63}& \textbf{92.99} &\textbf{93.11} &\textbf{92.16} \\
        SynCLR & StableRep & 99.99 & 96.05 & 97.72&\textbf{91.60}&99.95 &95.06&75.83&86.70 &99.57 &\textbf{81.35} &\textbf{98.64}&99.57& 72.66 & 92.72 & 78.51& 84.85 & 84.50 & 85.90& 90.07 \\
        \bottomrule
    \end{tabular}
    }
\end{table}
\begin{table}[tb]
    \caption{Classification accuracy for all combinations of backbone pre-training methods (rows) in detecting fake images from different generative models (columns). CLIP is the 32nd epoch of OpenCLIP, and $\text{CLIP}^{*}$ represents the weights of the 31st epoch. The combination of CLIP and $\text{CLIP}^{*}$ in the first row is the baseline.}
    \label{table:acc_ensemble}
    \centering
    \resizebox{\textwidth}{!}{%
    \begin{tabular}{ccp{3em}p{3em}p{3em}p{3.5em}p{3em}p{3em}p{3em}p{3em}p{3em}p{3em}p{3em}p{3em}p{3.5em}p{3em}p{3em}p{3.5em}p{3.5em}p{4em}p{3em}} 
    \toprule
        \multirow{2}{*}{Model} &
        \multicolumn{8}{c}{Generative Adversarial Networks} & 
        \multirow{2}{*}{\makecell[tl]{Deep\\fakes}} &
        \multicolumn{2}{c}{Low vision} &
        \multicolumn{2}{c}{Perceptual} &
        \multirow{2}{*}{Guided} &
        \multicolumn{2}{c}{LDM} &
        \multicolumn{2}{c}{Glide} &
        \multirow{2}{*}{\makecell[tl]{DALL-E}}&
        \multicolumn{1}{c}{Total}
        \\ \cmidrule(r){3-9} \cmidrule(r){11-14} \cmidrule{16-17} \cmidrule(r){18-19} \cmidrule{21-21} 
        & & Pro- GAN & Cycle- GAN & Big- GAN & Style- GAN2 & Gau- GAN & Star- GAN & Giga- GAN & & SITD &  SAN & CRN &  IMLE & & 200 steps & 200 w/CFG & 100 27 & 100 10 & &Avg. acc
        \\
        \midrule
        CLIP & $\text{CLIP}^{*}$ & 98.50 & 83.95 & 80.10& 58.25& 89.15 & 83.40& 56.30&72.95 & 67.00 & \textbf{58.00} & 61.95 & \textbf{92.60}& \textbf{82.05} & 81.80 & 62.50 & 71.95& 70.20 & 76.75 &74.86\\
        \midrule
        CLIP & DINOv2 & 99.90 & 82.10 & 90.60& 65.40& 97.05 & 87.80&\textbf{62.70}& 78.00 & 86.00 & 54.00 & 53.05 & 69.45& 75.05 & \textbf{89.50} & \textbf{69.50}& \textbf{73.45}& \textbf{71.90} & \textbf{78.85} &\textbf{76.91}\\
        CLIP & StableRep &  99.65& 81.55 & 71.15&67.60&92.40&80.55 &54.85&\textbf{80.75}&78.00&55.50 &55.95&67.40& 57.75 & 75.05 & 57.30& 60.40& 60.60 &63.55& 70.00 \\
        CLIP & SynCLR &99.90 &\textbf{88.60}  & \textbf{92.70}&62.40&\textbf{99.10}&\textbf{90.40}&58.25&78.05&\textbf{97.00}&\textbf{58.0}0 &\textbf{65.45}&88.55&56.25& 72.70 & 57.45& 62.50 & 62.65 & 67.90& 75.44\\
        DINOv2 & StableRep &  99.65& 77.80& 75.95&\textbf{71.25}&92.45 &74.80&57.10&78.05&92.50 &53.50 &52.30&55.05& 56.65& 81.30 & 62.20 & 66.25& 65.20 & 65.15 & 70.95\\
        DINOv2 & SynCLR & \textbf{99.95} & 85.15 &91.70&68.60&98.00 &87.70 &60.15&75.75 &93.50 &53.50&55.10&63.50& 58.25 & 80.45 & 61.10 & 70.60& 68.60 &70.90& 72.13 \\
        SynCLR & stableRep & 99.85 & 86.35 & 88.00&64.95&98.25 &86.30&55.55&77.60 &88.00 &56.00 &62.25&74.65& 52.15 & 69.05 & 55.50& 58.85 & 59.25 & 60.85 & 74.58\\
        \bottomrule
    \end{tabular}
    }
\end{table}

Tables~\ref{table:ap_ensemble} and~\ref{table:acc_ensemble} present the AP and classification accuracy, respectively, for all ensemble combinations (rows) in detecting synthetic images from different generative models (columns). For classification accuracy, the numbers shown are averaged over the real and fake classes for each generative model. Additionally, as a baseline for comparison, we use an ensemble of OpenCLIP weights from the 31st and 32nd epochs.

The ensemble of CLIP and SynCLR improves by $+$7.53 mAP and $+$0.58\% accuracy on average compared to the baseline. Additionally, compared to the ensemble of CLIP and DINOv2, the accuracy is slightly lower, but the AP is improved by approximately $+$3 mAP. These quantitative results demonstrate the potential of utilizing synthetic data-driven representations to enhance the performance of synthetic image detectors. However, similar to the evaluation of individual backbones, the detection performance for images generated by DMs remains relatively low. This indicates that a simple ensemble did not successfully combine the best features of the two backbones.

\section{Limitation}

Despite demonstrating the potential of synthetic data-driven representations for SID, this paper acknowledges its limitations. Firstly, UnivFD and its variants primarily use ViT as the backbone, and the publicly available pre-trained models for StableRep and SynCLR are only available for ViT. Therefore, all experiments in this study use ViT as the backbone architecture. However, to examine how synthetic data-driven general-purpose representations are influenced by different architectures, it is necessary to conduct evaluations using other architectures such as ResNet~\cite{he2016deep}. Additionally, the number of images used for pre-training each backbone is not exactly the same, so a comparison under completely identical conditions has not been achieved. Furthermore, both StableRep and SynCLR were pre-trained using images generated by DM, and it has not been verified whether symmetrical results would be obtained if they were pre-trained with GAN-generated images.
\section{Conclusion}
In this work, we studied the effectiveness of synthetic data-driven general-purpose representations for detecting fake images. Our comprehensive experiments across various datasets reveal the properties of synthetic data-driven representations and demonstrate their superiority over conventional representations learned from only real data in detecting generative models that were not used during pre-training. These findings highlight the potential of synthetic data-driven approaches in enhancing the robustness and accuracy of synthetic image detectors.

\section*{Acknowledgements}
We express our gratitude to Feng Qi for his valuable advice on the draft. This work was supported by JSPS KAKENHI Grant Numbers JP24H00748, JP24H00742, and JP21H05054.




%
%
\bibliographystyle{splncs04}
\bibliography{bib/article}
\end{document}